\documentclass[runningheads]{llncs}

\usepackage{amsmath,amsfonts,bm}

\def\eqref#1{equation~\ref{#1}}

\def\1{\bm{1}}

\DeclareMathAlphabet{\mathsfit}{\encodingdefault}{\sfdefault}{m}{sl}
\SetMathAlphabet{\mathsfit}{bold}{\encodingdefault}{\sfdefault}{bx}{n}

\usepackage[T1]{fontenc}
\usepackage{booktabs}

\usepackage{graphicx,verbatim}
\usepackage{hyperref}
\usepackage{url}
\usepackage{amsmath,amssymb,amsfonts}
\usepackage{algorithmic}
\usepackage{graphicx}
\usepackage{textcomp}
\usepackage{multirow}
\usepackage{tabu}
\usepackage{bbm}
\usepackage{diagbox}
\usepackage[english]{babel}
\usepackage{comment}
\usepackage{array}
\usepackage{amstext}
\usepackage{makecell}
\usepackage{caption}
\usepackage{tablefootnote}
\usepackage{babel}
\usepackage{booktabs} 
\usepackage{color}
\usepackage{amsmath}
\usepackage{wrapfig}
\usepackage{cleveref}
\usepackage{enumitem} 
\setlist[itemize]{noitemsep}

\begin{document}
\title{Pairwise-Constrained Implicit Functions for\\ 3D Human Heart Modeling}

\author{Hieu Le\inst{1} \and
Jingyi Xu\inst{2} \and
Nicolas Talabot\inst{1} \and
Jiancheng Yang\inst{1} \and
Pascal Fua\inst{1}
}
\authorrunning{Le et al.}
\institute{EPFL, Lausanne, Switzerland \and
Stony Brook University, New York, USA}
\titlerunning{Pairwise-Constrained Implicit Functions }

\maketitle              


\newif\ifdraft
\draftfalse
\drafttrue 

\definecolor{orange}{rgb}{1,0.5,0}
\definecolor{violet}{RGB}{70,0,170}
\definecolor{magenta}{RGB}{170,0,170}
\definecolor{dgreen}{RGB}{0,150,0}

\ifdraft
 \newcommand{\PF}[1]{{\color{red}{\bf PF: #1}}}
 \newcommand{\pf}[1]{{\color{red} #1}}
 \newcommand{\FS}[1]{{\color{blue}{\bf FS: #1}}}
 \newcommand{\fs}[1]{{\color{blue} #1}}
 \newcommand{\HL}[1]{{\color{orange} {\bf HL: #1}}}
 \newcommand{\hl}[1]{{\color{orange} #1}}
 \newcommand{\BG}[1]{{\color{olive}{\bf BG: #1}}}
 \newcommand{\bg}[1]{{\color{olive} #1}}
 \newcommand{\red}[1]{{\color{red}#1}}
 \newcommand{\todo}[1]{{\color{red}#1}}
 \newcommand{\TODO}[1]{\textbf{\color{red}[TODO: #1]}}
\else
 \newcommand{\PF}[1]{}
 \newcommand{\pf}[1]{#1}
 \newcommand{\FS}[1]{}
 \newcommand{\fs}[1]{#1}
 \newcommand{\hl}[1]{#1}
 \newcommand{\BG}[1]{}
 \newcommand{\bg}[1]{#1}
 \newcommand{\ME}[1]{}
  \newcommand{\me}[1]{#1}
  \newcommand{\TODO}[1]{}
  \newcommand{\todo}[1]{#1}
\fi

\def\ps{MLV}

\def\pz{Myocardium-LV}
\def\po{Left-Atrium}
\def\ptw{Left-Ventricle}
\def\pth{Right-Atrium}
\def\four{RV}

\newcommand{\parag}[1]{\vspace{-0mm}\paragraph{#1}}
\newcommand{\sparag}[1]{\subparagraph{#1}}
\newcommand*{\Scale}[2][4]{\scalebox{#1}{$#2$}}%
\newcommand*{\Resize}[2]{\resizebox{#1}{!}{$#2$}}%
\renewcommand{\floatpagefraction}{.99}

\newcommand{\bA}{\mathbf{A}}
\newcommand{\bC}{\mathbf{C}}
\newcommand{\bD}{\mathbf{D}}
\newcommand{\bH}{\mathbf{H}}
\newcommand{\bK}{\mathbf{K}}
\newcommand{\bP}{\mathbf{P}}
\newcommand{\bR}{\mathbf{R}}
\newcommand{\bX}{\mathbf{X}}
\newcommand{\bZ}{\mathbf{Z}}

\newcommand{\real}{\mathbb{R}}

\newcommand{\ba}{\mathbf{a}}
\newcommand{\bb}{\mathbf{b}}
\newcommand{\bc}{\mathbf{c}}

\newcommand{\f}{\mathbf{f}}
\newcommand{\bl}{\mathbf{l}}
\newcommand{\bI}{\mathbf{I}}

\newcommand{\bs}{\mathbf{s}}
\newcommand{\bt}{\mathbf{t}}
\newcommand{\bu}{\mathbf{u}}
\newcommand{\bw}{\mathbf{w}}
\newcommand{\bx}{\mathbf{x}}
\newcommand{\by}{\mathbf{y}}
\newcommand{\bz}{\mathbf{z}}

\newcommand{\radius}{\mathbf{r}}

\newcommand{\cF}{\mathcal F}
\newcommand{\fd}{\mathcal{F}_{d}}
\newcommand{\fz}{\mathcal{F}_{z}}

\newcommand{\OURS}[0]{\textbf{OURS}}
\newcommand{\FGSMU}[1]{\textbf{FGSM-U(#1)}}
\newcommand{\FGSMT}[1]{\textbf{FGSM-T(#1)}}
\newcommand{\FGSMUE}[1]{\textbf{FGSM-UE(#1)}}
\newcommand{\FGSMTE}[1]{\textbf{FGSM-TE(#1)}}

\newcommand{\colvecTwo}[2]{\ensuremath{
		\begin{bmatrix}{#1}	\\	{#2}	\end{bmatrix}
}}
\newcommand{\colvec}[3]{\ensuremath{
		\begin{bmatrix}{#1}	\\	{#2}	\\	{#3} \end{bmatrix}
}}
\newcommand{\colvecFour}[4]{\ensuremath{
		\begin{bmatrix}{#1}	\\	{#2}	\\	{#3} \\	{#4}	\end{bmatrix}
}}

\newcommand{\rowvecTwo}[2]{\ensuremath{
		\begin{bmatrix}{#1}	&	{#2}	\end{bmatrix}
}}
\newcommand{\rowvec}[3]{\ensuremath{
		\begin{bmatrix}{#1} &	{#2}	&	{#3} \end{bmatrix}
}}
\newcommand{\rowvecFour}[4]{\ensuremath{
		\begin{bmatrix}{#1}	&	{#2}	&	{#3} &	{#4}	\end{bmatrix}
}}

\newcommand{\tr}{^\intercal}


\begin{abstract}

Accurate 3D models of the human heart require not only correct outer surfaces but also realistic inner structures, such as the ventricles, atria, and myocardial layers. Approaches relying on implicit surfaces, such as signed distance functions (SDFs), are primarily designed for single watertight surfaces, making them ill-suited for multi-layered anatomical structures. They often produce gaps or overlaps in shared boundaries. Unsigned distance functions (UDFs) can model non-watertight geometries but are harder to optimize, while voxel-based methods are limited in resolution and struggle to produce smooth, anatomically realistic surfaces. We introduce a pairwise-constrained SDF approach that models the heart as a set of interdependent SDFs, each representing a distinct anatomical component. By enforcing proper contact between adjacent SDFs, we ensure that they form anatomically correct shared walls, preserving the internal structure of the heart and preventing overlaps, or unwanted gaps. Our method significantly improves inner structure accuracy over single-SDF, UDF-based, voxel-based, and segmentation-based reconstructions. We further demonstrate its generalizability by applying it to a vertebrae dataset, preventing unwanted contact between structures. 



\end{abstract}

\section{Introduction}

Accurate 3D modeling of the human heart is key to surgical planning, computational simulations, and disease diagnosis. However, the heart’s multi-chambered structure, with its ventricles, atria, and myocardial layers, poses a fundamental challenge: Accurate external reconstruction is not enough. Shared surfaces between individual components must be preserved, while preventing gaps, overlaps, and incorrect connections. 

Despite recent advances, existing methods still fail to do this. Voxel-based approaches such as nn-UNet~\cite{IsenseeEmail18,Isensee24} are resolution-limited, making smooth surface modeling difficult. Classic implicit functions, such as SDFs~\cite{Park19c}, yield higher-resolution reconstructions but are best for single watertight surfaces. Newer SDF-based approaches~\cite{Liu24,zhang22c} handle composite shapes but focus on overall shape reconstruction rather than delivering anatomical accuracy or meeting anatomical constraints. Unsigned distance functions (UDFs)~\cite{Chibane20b,rella24}
can model inner structures directly, but suffer from unstable optimization, which limits accuracy.

To address these challenges, we propose a novel pairwise-constrained implicit function approach that enforces proper anatomical contacts. We model the heart as a set of interdependent SDFs, each corresponding to a distinct anatomical component. Unlike previous methods that focus only on adjacency\cite{Gupta22} or avoiding interpenetration\cite{Karunratanakul20}, our approach ensures that adjacent components form contact regions with correct surface areas, accurately reflecting anatomical measurements. This matters because, when analyzing cardiac datasets, it has been observed that the ratio between the surface area of the contact region and the total surface area of paired components remain remarkably stable~\cite{Buckberg18,MedranoGracia14,Choy08}. 

This yields a previously unused constraint that we  leverage to improve reconstruction accuracy. To this end, we introduce a sampling-based verification strategy, where we randomly sample a large number of points in space to measure the actual contact ratio between components. By comparing this empirical ratio to the expected anatomical values from the training set, we iteratively adjust the implicit function representation, ensuring that the reconstructed heart maintains both topological and anatomical correctness.

We validate our method on cardiac MRI datasets, demonstrating that it significantly improves inner structure accuracy over single-SDF, UDF-based, and voxel-based approaches. To demonstrate its generalizability, we also apply it to a vertebrae dataset, where it successfully maintains proper spacing between adjacent vertebrae - a very different constraint from the heart. This highlights the promise of pairwise-constrained implicit functions for achieving high-fidelity, multi-component medical 3D modeling.

\section{Related work}

Modeling the internal structure of the human heart requires preserving anatomical relationships between adjacent chambers while ensuring accurate reconstruction of shared surfaces. Some segmentation-based methods enforce spatial consistency using topology-aware losses~\cite{Hu19b,Hu21b} or multi-category spatial constraints~\cite{Gupta22}, but their voxelized nature limits resolution and smoothness. Mesh-based approaches like Voxel2Mesh~\cite{Wickramasinghe20} refine outputs into meshes but struggle with complex anatomical details due to their reliance on simple geometric templates.

Classic implicit functions, such as SDFs~\cite{Park19c}, yield high-resolution reconstructions but were only intended for single, watertight surfaces. Multi-SDF approaches such as MODIF~\cite{Liu24}, IPM\cite{zhang22c}, or SOMH~\cite{Verhulsdonk24} extend this to more complex structures but do not explicitly enforce inter-object constraints needed for accurate shared surface reconstruction. Our method addresses this limitation by enforcing anatomical contact regions with precise surface area constraints, ensuring structural integrity. UDFs~\cite{Chibane20b} provide an alternative for modeling non-watertight surfaces but suffer from unstable optimization~\cite{rella24,Stella24}, making them unreliable for medical applications. Our experiments show that, when applied to heart modeling, UDF-based methods often generate incomplete or missing surfaces, further limiting their usability.

\section{Method}
\label{sec:method}

Our method ensures that adjacent structures maintain proper contacts of the correct extent. In heart reconstruction, it preserves the correct amount of the shared surface area between chambers~\cite{Buckberg18}. These shared surface ratios are remarkably stable and can be reliably derived from the training set. For example, the wall between the left ventricle and its myocardium consistently accounts for 27\% of their combined surface area and analysis of 120 training instances only shows deviations of less than 2\% in the most extreme cases. Our approach enforces these ratios, preventing unnatural gaps or overlaps. Similarly, when applying our approach to  the reconstruction of healthy human spines, where adjacent vertebrae must maintain a minimum gap of at least 1 mm~\cite{Little05}, we introduce separation constraints. 

Our framework handles both scenarios equally well, even though they involve very different constraints. For contact ratios, the primary difficulty lies in ensuring that objects interact without penetrating each other, which requires precise alignment and continuous adjustment of surface boundaries. For minimum distance constraints, detecting violations necessitates global checks to ensure that no parts of the objects come too close from each other. Addressing these two distinct sets of requirements using traditional modeling frameworks can be cumbersome and often involves separate, specialized approaches. In contrast, our approach handles both cases in a consistent, unified manner.



Our method relies on SDFs~\cite{Park19c}, which have emerged as a powerful model to learn continuous representations of 3D shapes.  They allow detailed reconstructions of object instances as well as meaningful interpolations between them.  Given an object, a signed distance function outputs the point's distance to the closest object surface: $f(\bz,\bx) = s: \bx\in\mathbb{R}^3, s\in\mathbb{R} \; , \label{eq:sdf}$ where $\bz$ is a latent vector that parameterizes the surface. In the remainder of this section, we first introduce our approach to enforcing contact ratio constraint in the heart. We then show how it can be extended to enforcing the minimum distance constraint between vertebrae. 


\subsection{Enforcing Correct Shared Surface Areas}
\label{sec:contact}

Given objects represented by their SDFs and a prior indicating a desired contact ratio (\%)  between them, our goal is to refine their 3D shapes in such a way that they do not intersect and touch each other within a small tolerance of the desired contact ratio. We do so by randomly sampling points and identifying ones lying in critical regions and adjusting the distances of all implicit shapes to those points to achieve the desired contact ratio. The key questions we need to address are \textit{how do we select these points},  \textit{how many do we need}, and  \textit{how to use them to adjust the shapes}. 

In our experiment, we reconstruct simultaneously five components of the human heart: The four ventricles and the myocardium of the left-ventricle, with the contact ratios between all pairs being given. However, for notational simplicity, in the remainder of this section we describe our approach in terms of only two components with SDFs, $f_A$ and $f_B$, whose network weights are frozen. We refine the shapes by minimizing the losses defined below with respect to the corresponding latent vectors, $\ba$ and $\bb$.  This generalizes naturally to all five.

\parag{\textbf{Mining Topologically Meaningful Points.}}

To enforce shared surface constraints, we introduce loss functions for three key point sets obtained by random sampling: 
(1) $\mathcal{A}_{\textrm{contact}}$ – points that should be in contact with both objects, 
(2) $\mathcal{A}_{\textrm{intersecting}}$ – points that sit inside both objects, and 
(3) $\mathcal{A}_{\textrm{ncontact}}$ – points that should not be in contact areas.  First, for two \textit{explicit} surfaces $A,B$, the contact ratio between $A$ and $B$ is defined as 
\begin{equation}
 P_{A,B}=\frac{\textrm{Area}(S_{AB})}{\textrm{Area}(S_{A})+ \textrm{Area}(S_{B})} \; ,
 \label{eq:real}
\end{equation}
where $S_{AB}$ represents the partial surface of object $A$ that is in close proximity, \textit{i.e.}, contact distance, to object $B$, and $S_{A}$ and $S_{B}$  refer to the entire surfaces of objects $A$ and $B$, respectively. For implicit objects, this can be approximated as
\begin{equation}
    P'_{A,B} = \text{\footnotesize $ \frac{\sum_{i=1}^N\mathbbm{1}(|f_{A}(\ba,\bx_i)|<\epsilon) \cdot \mathbbm{1}(|f_{B}(\bb,\bx_i)|<\epsilon)}{\sum_{i=1}^N\mathbbm{1}(|f_{A}(\ba,\bx_i)|<\epsilon) + \sum_{i=1}^N\mathbbm{1}(|f_{B}(\bb,\bx_i)|<\epsilon)}, $}
\label{eq:random}
\end{equation}
where $\{\bx_i: i \in(1,N)\}$ is a set of $N$ uniformly sampled random points, $\epsilon$ is a small threshold indicating the contact distance, and $\mathbbm{1}(\cdot)$ is the indicator function that returns the value 1 if its statement is true and 0 otherwise. 

Our goal is to adjust $\ba$ and $\bb$ so that the estimated contact surface $P'_{A,B}$ aligns with the expected value $P_{A,B}$. To achieve this, we modify the numerator of \cref*{eq:random}, which represents the contact surface. While the denominator, which reflects the combined surface areas of the objects, could also be altered, it tends to remain stable during optimization as the overall object sizes are largely determined by the initial segmentation inputs. 
Thus, the expected number of points in the contact region should be
\begin{equation}
N_{\textrm{contact}} = \textrm{P}_{A,B} \times \left(\sum_{i=1}^N\mathbbm{1}(|f_{A}(\ba,\bx_i)|<\epsilon) + \sum_{i=1}^N\mathbbm{1}(|f_{B}(\bb,\bx_i)|<\epsilon)\right),
\end{equation}
where $\textrm{P}_{A,B}$ is the expected proportion of shared surface area between the two components, which can be easily computed from the training shapes. To enforce this expected number of contact points, we define three sets of points for optimization. 1) $\mathcal{A}_{\textrm{contact}}$ : The top $N_{\textrm{contact}}$ points with closest summed distance to both objects. 2) $\mathcal{A}_{\textrm{ncontact}}$ : All points that are in close proximity to both $f_{A}$ and $f_{B}$ but are not included in $\mathcal{A}_{\textrm{contact}}$, and 3) $\mathcal{A}_{\textrm{intersecting}}$: all points that are within both objects: $\mathcal{A}_{\textrm{intersecting}} = \{\bx \mid f_A(\ba,\bx)<0 \land f_B(\bb,\bx) <0 \}$. These three sets, which are periodically resampled, allow us to define the following loss functions whose minimization ensures correct contacts, while preventing overlap.


\parag{\textbf{Contact Ratio Loss.}}

To ensure the correct shared surface, we define
\begin{small}
\begin{equation}
    \mathcal{L}_{\textrm{contact}} = \!\!\!\!\! \sum_{\bx \in \mathcal{A_{\textrm{contact}}}}  \!\!\! \!\!\! | f_A(\ba,\bx) + f_B(\bb,\bx)| \mbox{ and }
    \mathcal{L}_{\textrm{ncontact}}  = \!\!\! \!\!\! \!  \sum_{\bx \in \mathcal{A_{\textrm{ncontact}}}} \!\!\!\!\!\!  -(f_A(\ba,\bx) + f_B(\bb,\bx)) .
\end{equation}
\end{small}
%
%
Minimizing $\mathcal{L}_{\textrm{contact}}$ encourages contact points to lie equally between both surfaces without overlap, reaching zero when $f_A(\ba,\bx) = -f_B(\bb,\bx)$. Minimizing $\mathcal{L}_{\textrm{ncontact}}$ prevents excessive contact.

\parag{\textbf{Intersection and data Losses.}}
To prevent intersections and ensure that objects remain faithful to their segmentation, we define
\begin{small}
\begin{align}
\mathcal{L}_{\textrm{intersecting}} &= \sum_{\bx \in \mathcal{A}_{\textrm{intersecting}}} |f_A(\ba,\bx)| + |f_B(\bb,\bx)| \; , \\
\mathcal{L}_{\textrm{data}} & = \sum_{\bx \in \bX_A} |f_A(\ba,\bx) - s_{\bx}| + \sum_{\bx \in \bX_B} |f_B(\bb,\bx) - s_{\bx}| \; ,
\end{align}
\end{small}
where $\bX_A$ and  $\bX_B$ represent 500'000 spatial locations per object and are sampled more aggressively (90\%) near the object surfaces.



\parag{\textbf{Optimization.}}

We reconstruct both objects by minimizing
\begin{small}
\begin{equation}
    \mathcal{L} = \lambda_1 \mathcal{L}_{\textrm{intersecting}} + \lambda_2 \mathcal{L}_{\textrm{contact}} + \lambda_3 \mathcal{L}_{\textrm{ncontact}} + \lambda_4 \mathcal{L}_{\textrm{data}} \; .
\end{equation}
\end{small}
To adapt to shape changes, we resample 300K points every 10 iterations to update topologically meaningful points.

\subsection{Enforcing Minimum Distance Constraints}
\label{sec:distance}

Our method can also ensure a minimum separation between adjacent structures when direct contact is not anatomically valid, such as between vertebrae in the spine. To achieve this, we identify points where the sum of distances to both surfaces falls below a given threshold $d$, forming the violation set $\mathcal{A}_{\textrm{violation}}= \{\bx \mid f_A(\ba,\bx) + f_B(\bb,\bx) < d\}.$ We then define the composite loss
\begin{small}
\begin{equation}
\mathcal{L}_{\textrm{min-distance}} =   \lambda_1 \!  \!  \! \sum_{\bx \in \mathcal{A}_{\textrm{violation}}}  \!  \!  \! \max(0, d - (f_A(\ba,\bx) + f_B(\bb,\bx))) + \lambda_2 \mathcal{L}_{\textrm{data}}  \; ,
\end{equation}
\end{small}
whose minimization fits the data while pushing the surfaces apart.


\section{Experiments}

We validate our method by reconstructing human heart and lumbar spines. We jointly reconstruct their components while enforcing proper constraints between them by fitting deep implicit functions~\cite{Park19c} to segmentation outputs from nnU-Net~\cite{IsenseeEmail18}.  We compare against several baselines: (1) a standard SDF method~\cite{Park19c} that fits each part individually, (2) a multi-SDF method based on template deformation~\cite{zhang22c}, (3) a multi-SDF approach that employs a single perceptron to implicitly learn part compatibility~\cite{Verhulsdonk24}, (4) a simple UDF-based method~\cite{Chibane20b} for modeling non-watertight shapes, and (5) a recent, state-of-the-art UDF-like representation for modeling open surfaces~\cite{rella24}. 

\begin{figure*}[t]
    \begin{center}
    \includegraphics[width=0.9\linewidth]{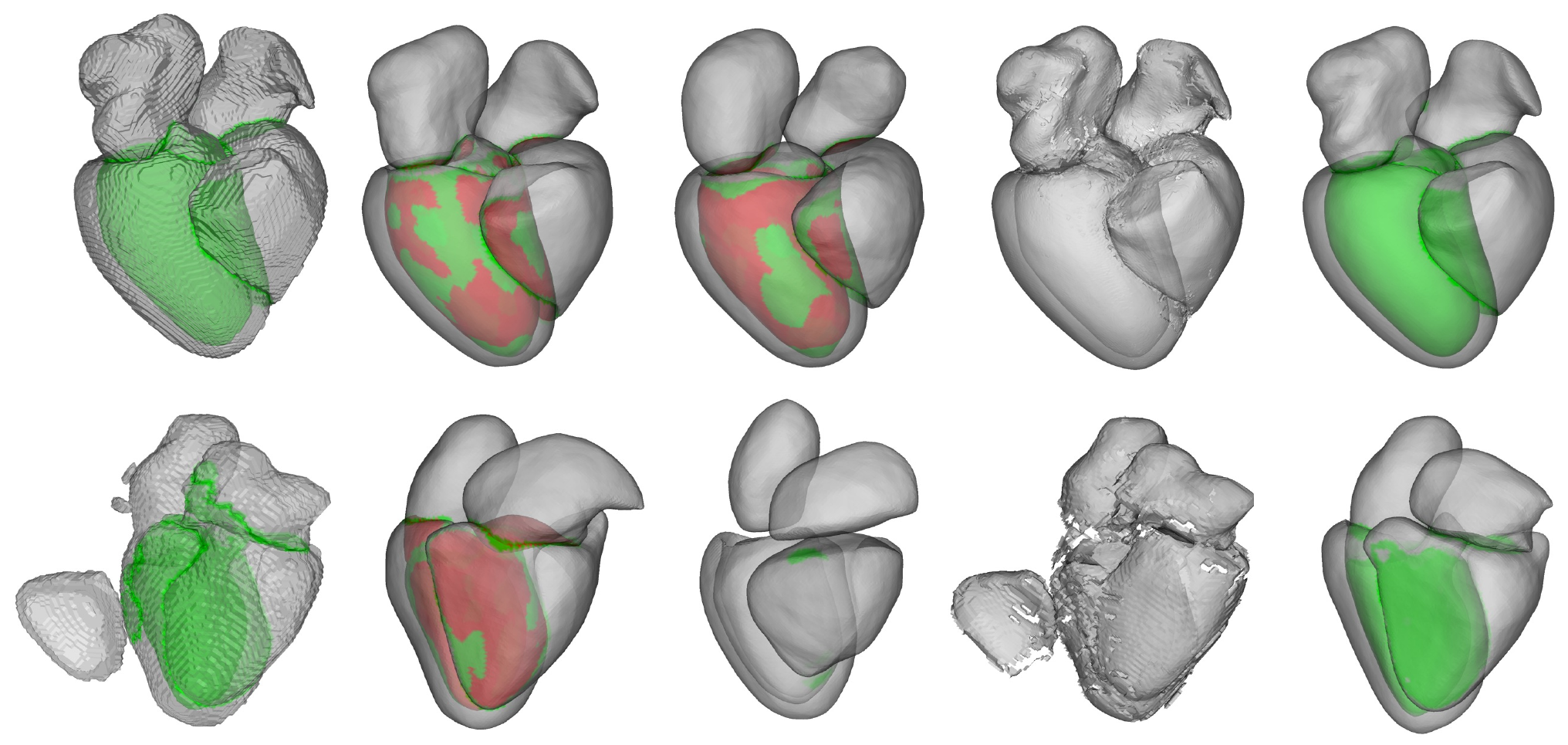}
    \footnotesize
    \makebox[0.17\linewidth]{(a)nnU-net\cite{Isensee24}}
    \makebox[0.17\linewidth]{(b)DeepSDF\cite{Park19c}}
    \makebox[0.17\linewidth]{(c)SOMH\cite{Verhulsdonk24}}
    \makebox[0.17\linewidth]{(d)NVF\cite{rella24}}
    \makebox[0.17\linewidth]{(e)Our}
    
    \end{center}
    \vspace{-5mm}
    \caption{The two rows show results from an in-distribution (ID) and an out-of-distribution (OOD) heart sample. We show transparent meshes to highlight contact areas. Red indicates penetration, while green denotes proper shared surfaces.}
    \label{fig:compositeheart}
    \vspace{-5mm}
  \end{figure*}
 
\begin{table}[!t]
    \centering
    \vspace{-3pt}
    \setlength{\tabcolsep}{10pt}
    \caption{\textbf{3D Heart Reconstruction.} We compare methods based on Chamfer distance ($\times 10^{4}$), area differences (Area), and normal consistency (NC) relative to the ground-truth meshes.}
    \label{tab:heart_inner}
    \begin{tabular}{lcccccc}
    \multirow{2}{*}{Method} & \multicolumn{3}{c}{In-Distribution} & \multicolumn{3}{c}{In-House Testing Data} \\
    \cmidrule(lr){2-4} \cmidrule(lr){5-7}
    & CD & Area & NC  
      & CD & Area & NC  \\
    \midrule
    nnU-Net\cite{Isensee24}  & \textbf{1.5} & 0.89 & 68.2     &5.0  &1.7  & 87.6  \\
    SDF~\cite{Park19c}       & 1.6    & 2.66       & 98.8     &5.0  &2.0  & \textbf{96.8} \\
    IPM~\cite{zhang22c}      & 10.7   & 4.14       & 98.7     & 10.4  &4.4  & 95.2 \\
    SOMH~\cite{Verhulsdonk24}& 1.9    & 0.81       & 99.2     & 7.2   &1.9  & \textbf{96.8} \\
    UDF~\cite{Chibane20b}    & \textbf{1.5}  & 0.40& 89.0      &40.6  & 1.8  & 73.4  \\
    NVF~\cite{rella24}       & 1.6    & 0.49   & 99.3         &4.8  &1.7 & 81.3  \\
    \midrule
    Ours   & \textbf{1.5} & \textbf{0.13} & \textbf{99.4}     &\textbf{4.1}  &\textbf{1.2}  & 95.3 \\
    \bottomrule
    \end{tabular}
   \vspace{-5pt}
\end{table}

\parag{3D Heart Reconstruction.} We use our method to reconstruct five heart components: left ventricle (LV), myocardium of the left ventricle (M-LV), left atrium (LA), right atrium (RA), and right ventricle (RV). We train an SDF auto-decoder~\cite{Park19c} on a public dataset of 120 whole-heart models~\cite{zhuang19}, setting aside 20 validation samples used for choosing hyper-parameters. Contact ratios between components are precomputed from the training data and used as constraints during reconstruction. The nnU-Net segmentation model is trained on 15 cardiac images from the same dataset, with only 20 images publicly available.
We evaluate our method on two test sets: (1) a public set of five cardiac images from the same distribution as the training data, and (2) an in-house dataset of 10 cardiac images from a local hospital, serving as an out-of-distribution test set due to significant differences in image quality. The ground-truth meshes are produced by applying Marching-Cube\cite{Lorensen87} and Laplacian smoothing on the annotated segmentations, similar to \cite{Kong21a}. We assess reconstruction accuracy using Chamfer distance (CD), surface area distance (Area), and normal consistency (NC).  

\begin{table}[h!]
\centering
\vspace{-3mm}
\caption{\textbf{Contact Ratio Statistics.} Our method can output shapes with similar contact ratios observed from the training set. }
\label{tab:cr}
\begin{tabular}{l|c|c|c|c}
Pairs           &LV-MLV & MLV-RV  & LV-RA  & RV-LA  \\
\midrule
Training Data      & $27.0\%\pm 0.06$ &  $8.6\%\pm 0.03$ & $5.5\%\pm 0.01$ & $7.6\% \pm0.03$     \\
Our output (Val)         & $27.1\%\pm 0.01$ &  $8.8\%\pm 0.01$ & $5.4\%\pm 0.03$ & $7.3\% \pm0.01$     \\
Our output (Test)         & $25.5\%\pm 0.02$ &  $10.0\%\pm 0.08$ & $5.0\%\pm 0.03$ & $5.6\% \pm0.08$     \\
\bottomrule
\end{tabular}
\vspace{-3mm}
\end{table}

\begin{figure*}[]
    \begin{center}
    \includegraphics[width=\linewidth]{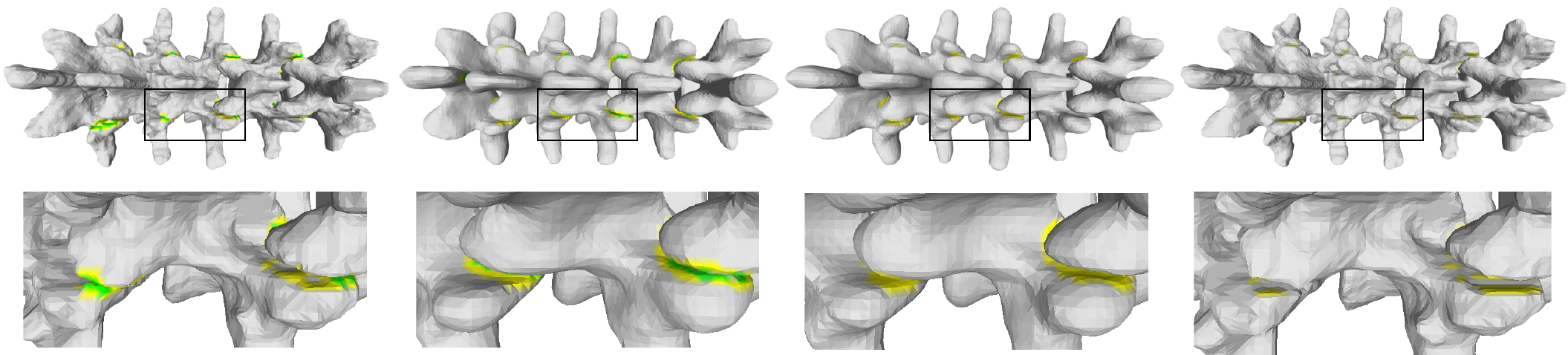}
    \makebox[0.24\linewidth]{(a) nnU-net }
    \makebox[0.24\linewidth]{(b) SDF\cite{Park19c} }
    \makebox[0.24\linewidth]{(c) Ours }
    \makebox[0.24\linewidth]{(d) Ground-truth }
    \end{center}
   \vspace{-5mm}
    \caption{\textbf{Verterbrae reconstruction.} The second row shows zoomed-in crops of the first row. Yellow indicates close but allowable proximity, while green marks undesired contact. 
    }
    \vspace{-2mm}
    \label{fig:spine}
    \end{figure*}

 As shown in Tab.~\ref{tab:heart_inner}, our method outperforms existing approaches across both datasets. For in-distribution data, it achieves the lowest CD (1.5, tied with nnU-Net and UDF), the lowest area difference (0.13), and the highest NC (99.4\%). Area difference is crucial as it indicates the absence of double or missing surfaces, and our method performs best in this regard. On in-house data, we maintain the best CD and area difference while achieving strong NC. Fig.~\ref{fig:compositeheart} further illustrates these results. For the OOD sample in the bottom row, nnU-Net struggles with severe misalignments and fragmented structures, as seen in the second row. Individually fitting SDFs reduces Chamfer distance by imposing priors on noisy data but introduces major topological errors. Our method further lowers CD while preserving structural integrity, outputing shapes with very similar contact ratios to ones in the training set, as can be seen in Tab.~\ref{tab:cr}. While SOMH\cite{Verhulsdonk24} achieves high NC, this is primarily due to overly smoothed outputs, as it fails to find an embedding that properly fits the noisy input segmentation. UDF-based methods~\cite{Chibane20b,rella24} do not produce any surface penetration by design, but their reconstructed meshes exhibit many missing holes due to unstable optimization.


\begin{table*}[th]
    \centering
       \vspace{-5mm}
    \caption{{\bf Spine reconstruction.} Comparison of three methods by constraint violations (area \& point count) and Chamfer distance (CD) across vertebrae pairs, and overall averages.}
    \label{tab:spine}
    \begin{tabular}{lccccccccc}
    \multirow{2}{*}{Method}& \multicolumn{2}{c}{L1 - L2} & \multicolumn{2}{c}{L2 - L3} &\multicolumn{2}{c}{L3 - L4}&\multicolumn{2}{c}{L4 - L5} & All \\
    \cmidrule(lr){2-3}  
    \cmidrule(lr){4-5}
    \cmidrule(lr){6-7}
    \cmidrule(lr){8-9}
    \cmidrule(lr){10-10}
     & px$^2$ & N  & px$^2$ & N & px$^2$ & N &px$^2$ & N  & CD ($\times10^3$) \\
    \cmidrule(lr){1-1}  
    \cmidrule(lr){2-3}  
    \cmidrule(lr){4-5}
    \cmidrule(lr){6-7}
    \cmidrule(lr){8-9}
    \cmidrule(lr){10-10}
    nn-Unet        & 261.4 & 123.2 & 527.2  & 237.3 & 644.6 & 294.3 & 761.1 & 358.9 & 0.4   \\       
    SDF     & 315.8 & 154.2 & 309.8  & 166.1& 416.4 & 227.7 & 503.3 & 279.4 & 0.4   \\                                 
    Ours   & \textbf{0.0} & \textbf{0.8} & \textbf{0.5}  & \textbf{2.3}& \textbf{0.5 }& \textbf{2.9 }& \textbf{1.2 }& \textbf{3.9 }& \textbf{0.3}   \\                   
    \bottomrule
    \end{tabular}
\end{table*}

\vspace{-3mm}
\parag{3D Lumbar Spine Segmentation.}We also use our method to reconstruct vertebrae in a dataset containing 460 CT images of the five lower vertebrae of the human spine (L1-L5)~\cite{Wasserthal22}. Of these, 80\% are used to train both the nnU-Net and the latent implicit models, while 10\% samples are reserved for testing and 10\% are for validation. In the entire dataset, each pair of adjacent vertebrae exhibits a minimum gap of 1 pixel, which is the constraint we enforce during reconstruction. In Tab.~\ref{tab:spine}, we report topological errors measured by the number of contact vertices (N) and the contact surface areas (px$^2$), along with reconstruction accuracy. Our approach yields reconstructions with almost no constraint violations. Note that enforcing the minimum gap constraint is particularly challenging because there are only small volumes surrounding each vertebrae joints that are topologically relevant, as can be seen in the highlighted areas in Fig.~\ref{fig:spine}. 




\begin{table}[h!]
\centering
\vspace{-3mm}
\caption{\textbf{Ablation study.} Performance of our method when reconstructing a left-ventricle and its myocardium when removing specific loss components.}
\label{tab:ablation}
\begin{tabular}{l|cccc|c|c}
Method           & w/o $L_{inter}$ & w/o $L_{contact}$ & w/o $L_{non-contact}$ & w/o $L_{data}$ &All & GT \\
\midrule
Penetration      & 5\%             & 0\%               & 0\%                   & 0\%     &0\%       & 0\% \\
Contact          & 17.2\%          & 13\%              & 29\%                  & 26\%    & 27\%      & 27\% \\
CD ($\times10^4$) & 3.8             & 3.7               & 3.5                   & 33.2   &  3.1      & 0 \\
\bottomrule
\end{tabular}
\vspace{-3mm}
\end{table}

\parag{Ablation.} 
 \vspace{-3mm}
We gauge the effect of each loss function of Section~\ref{sec:contact}  by omitting each one in turn when reconstructing a left-ventricle / myocardium pair in the out-of-distribution test set. As can be seen by comparing the ``w/o'' columns of Tab.~\ref{tab:ablation} against the one that shows the results when using ``all'' losses, they all contribute to the accuracy of the final result.

\section{Conclusion}

We have introduced a novel method to  enforcing topological consistency  constraints between multiple 3D objects modeled in terms of implicit surfaces. In the heart reconstruction scenario, our method maintains a precise contact ratio while preventing interpenetrations. In the lumbar spine reconstruction case, we ensure that adjacent vertebrae does not touch each other. This is achieved via randomized sampling of well chosen object locations. Future work will explore more complex scenarios, such as enforcing the constraints only at specific locations and not at others where they do not apply due to various pathologies. This will be turned into a powerful diagnostic tools to detect the extent of these pathologies.

\clearpage

\bibliographystyle{splncs04}
\bibliography{bib/string,bib/vision,bib/graphics,bib/biomed,bib/med}

\end{document}